\documentclass[runningheads]{llncs}
\usepackage{graphicx}
\usepackage{amsmath,amssymb} 
\usepackage{color}
\usepackage{hyperref}

\begin{document}
\title{Cross and Learn: Cross-Modal Self-Supervision}

\titlerunning{Cross and Learn}
%
\author{Nawid Sayed\inst{1} \and
Biagio Brattoli\inst{2} \and
Bj{\"o}rn Ommer\inst{2}}
%
\authorrunning{N. Sayed and B. Brattoli and B. Ommer}
%

\institute{Heidelberg University, HCI / IWR, Germany \\
	\inst{1}\email{nawid.sayed42@gmail.com} \\
	\inst{2}\email{\{firstname.lastname\}@iwr.uni-heidelberg.de}}
\maketitle              
\begin{abstract}
In this paper we present a self-supervised method for representation learning utilizing two different modalities. Based on the observation that cross-modal information has a high semantic meaning we propose a method to effectively exploit this signal. For our approach we utilize video data since it is available on a large scale and provides easily accessible modalities given by RGB and optical flow. We demonstrate state-of-the-art performance on highly contested action recognition datasets in the context of self-supervised learning. We show that our feature representation also transfers to other tasks and conduct extensive ablation studies to validate our core contributions. Code and model can be found at \url{https://github.com/nawidsayed/Cross-and-Learn}.
\end{abstract}

\section{Introduction}
\begin{figure}[t]
	\centering
	\includegraphics[width=\linewidth]{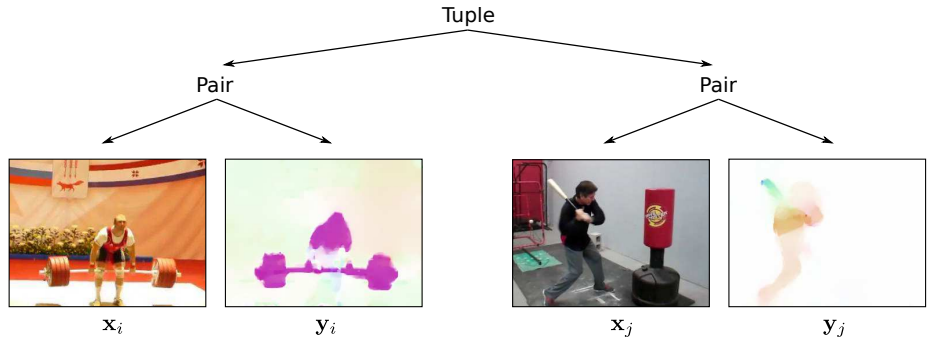}
	\caption{A tuple of pairs obtained from UCF-101 using RGB and optical flow as modalities. The cross-modal information (barbell, bat, person) has a much higher semantic meaning than the modality specific information (background, camera movement, noise)}
	\label{fig:image_pair_tuple}
\end{figure}
In the last decade, Convolutional Neural Networks (CNNs) have shown state of the art accuracy on a variety of visual recognition tasks such as image classification \cite{alexnet}, object detection \cite{fastrcnn} and action recognition \cite{twostream14}. The success of CNNs is based on \textit{supervised learning} which heavily relies on large manually annotated datasets. These are costly to obtain, however unannotated data in form of images and videos have become easily available on a very large scale. Recently this encouraged the investigation of \textit{self-supervised learning} approaches which do not require semantic annotations for the data. Here the general procedure involves pretraining a network on a surrogate task which requires semantic understanding in order to be solved. Although the gap is closing, self-supervised approaches usually cannot compete with feature representations obtained by supervised pretraining on large scale datasets such as ImageNet \cite{imagenet} or Kinetics \cite{kinetics}.

In this paper, we use cross-modal information as an alternative source of supervision and propose a new method to effectively exploit mutual information in order to train powerful feature representations for both modalities. The main motivation of our approach is derived from the following observation: Information shared across modalities has a much higher semantic meaning compared to information which is modality specific. We showcase this point in Fig. \ref{fig:image_pair_tuple} where we present a tuple of pairs obtained from an action recognition video dataset. We can see that cross-modal information such as the barbell or the baseball bat provide good clues to identify the action. On the other hand modality specific information such as the background or camera motion do not help to identify the action.

Feature representations which are sensitive to cross-modal information and invariant to modality specific content are therefore desirable. The latter condition is fulfilled if the feature representations of a pair are similar to each other. The former condition is fulfilled if the feature representations are also dissimilar across different pairs. To achieve that we utilize a trainable two stream architecture with one network per modality similar to \cite{twostream14} and propose two different loss contributions which guide the networks to learn the desired feature representations. 

For our method we require paired data from different modalities on a large scale. We therefore apply our method to video data which provide very easily accessible modalities RGB and optical flow in practically unlimited quantity. Our choice for the modalities is also motivated by past work where it has been shown that RGB and optical flow complement each other in the context of action recognition \cite{twostream14}. 

In order to demonstrate the effectiveness of our cross-modal pretraining, we conduct extensive experimental validation and a complete ablation study on our design choices. Our method significantly outperforms state-of-the-art unsupervised approaches on the two highly contested action recognition datasets UCF-101 \cite{ucf} and HMDB-51 \cite{hmdb} while pretraining for only \textit{6h GPU time} on a single NVIDIA Titan X (Pascal). We also show the transferability of our feature representation to object classification and detection by achieving competitive results on the PASCAL VOC 2007 benchmark.

\section{Related Work}
In recent years, unsupervised deep learning has been an hot topic for the research community due to the abundance of unannotated data. One of the first approaches was the Auto-Encoder \cite{ae,vae} which learns to encode an image in a lower dimensional features space. The obtained feature representation would mostly focus on low level statistics instead of capturing the semantic content of the input.

A significant step forward in deep unsupervised learning was made by \cite{contextpred,egomotion,wang2015} which introduced the paradigm of self-supervision in the context of deep learning. This family of methods exploits the inherent structure of data in order to train a CNN through a surrogate task. The most recent approaches can generally be divided into two groups, those which make use of static images and those which make use of videos. 

As for static images, there are several approaches exploiting color information \cite{colorization16,colorization17} and spatial context \cite{contextpred,jigsaw,jigsaw++}. For the latter an image is tiled into patches and their relative position is exploited as a supervisory signal in order to obtain a good feature representation. These approaches have shown reliable results on the task of object detection and surface normal estimation. There is also a variety of works which specifically aim to learn reliable posture representations without supervision \cite{cliqueCNN,posefromaction,timo,artsiom,brattoli}. 

Approaches which use video data mainly exploit temporal context as source of supervision \cite{wang2015,shufflealearn,brattoli,buechler,miguelpatrick}. A common idea is to shuffle frames of a video clip and use a network to reconstruct or verify the correct chronological order \cite{shufflealearn,oddoneout,opn}. The obtained feature representations generalize well to action recognition. Closely related to our approach is the work of S. Purushwalkam and A. Gupta \cite{posefromaction}. They use RGB and optical flow data obtained from videos and learn a feature representation which transfers to the task of pose estimation. Although they achieve good results for action recognition, our method exploits cross-modal information in RGB and optical flow data much more effectively.

There is a large body of work which demonstrates the effectiveness of utilizing multimodal data in order to solve a variety of tasks. For example \cite{twostream14,twostream16} show that RGB and optical flow are complementary modalities in the context of action recognition. 

S.Gupta et al. \cite{crossmodaldist} train a feature representation for depth data by distilling the knowledge from a teacher network pretrained on ImageNet \cite{imagenet}. The work of \cite{embedding} proposes cross-modal ranking in conjunction with structure preserving constrains in order to learn a joint feature representation between image and text data. Both works show successful cross-modal supervision but rely on annotated data, which is not required for our approach.

The work of \cite{multimodalae} utilizes a bimodal deep autoencoder in order to learn a shared feature representation for the audio and visual stream of video data. Similarly \cite{corrrnn} extends this work to temporal feature representations.
\section{Approach}
\label{sec:method}
\begin{figure}[t]
	\centering
	\includegraphics[width=\linewidth]{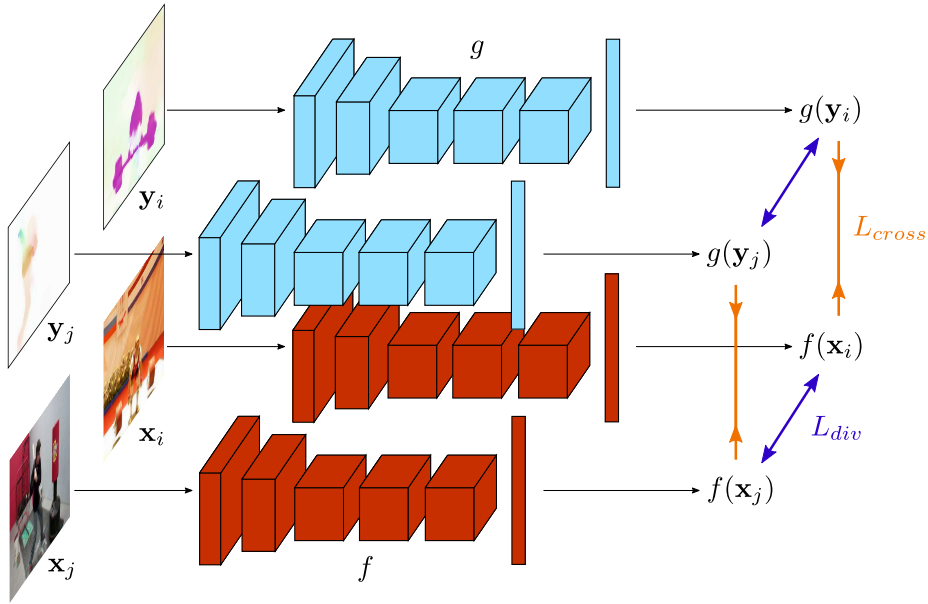}
	\caption{Visualization of our model. Networks of same color share weights and are denoted by $f$ (red, dark) and $g$ (blue, bright) respectively. $f$ and $g$ are trained so that the cross-modal distance (vertical) decreases while the distance between different pairs (horizontal) increases}
	\label{fig:image_method}
\end{figure}
In this section we present an efficient method for representation learning utilizing two different modalities. Our goal is to obtain feature representations which are sensitive to cross-modal information while being invariant to modality specific content. As already explained in the introduction, these conditions are fulfilled by feature representations which are similar for a pair and dissimilar across different pairs. To achieve the former we propose a cross-modal loss $L_{cross}$ and to achieve the latter we utilize a diversity loss $L_{div}$, both of which act directly in feature space thus promising better training signals.

Our method requires paired data from two different modalities $\mathbf{x}\in X$ and $\mathbf{y}\in Y$, which is available in most use cases i.e. RGB and optical flow. We utilize a two-stream architecture with trainable CNNs in order to obtain our feature representations $f(\mathbf{x})$ and $g(\mathbf{y})$. With exception of the first layer, the networks share the same architecture but do not share weights. To calculate both loss contributions we need a tuple of pairs $\mathbf{x}_{i}, \mathbf{y}_{i}$ and $\mathbf{x}_{j}, \mathbf{y}_{j}$ from our dataset. A visualization of our method is provided in Fig. \ref{fig:image_method}.

\subsubsection{Cross-Modal Loss.}
In order to enforce cross-modal similarity between $f$ and $g$ we enforce the feature representations of a pair to be close in feature space via some distance $\text{d}$. Solving this task requires the networks to ignore information which is only present in either $\mathbf{x}$ or $\mathbf{y}$
\begin{equation}
\label{eq:loss_cross}
L_{cross}(\mathbf{x}_i,\mathbf{y}_i,\mathbf{x}_j,\mathbf{y}_j) = \frac{1}{2}\left[\text{d}(f(\mathbf{x}_i), g(\mathbf{y}_i)) + 
\text{d}(f(\mathbf{x}_j), g(\mathbf{y}_j))\right].
\end{equation}	
We utilize the bounded cosine distance for $\text{d}$, which is given by 
\begin{equation}
\label{eq:cos_dist}
\text{d}(\mathbf{a},\mathbf{b}) = 1-\frac{\mathbf{a} \cdot \mathbf{b}}
{\vert \mathbf{a} \vert\cdot\vert \mathbf{b} \vert} \leq 2.
\end{equation}	
We prevent singularities during training by adding a small $\epsilon$ into the square-root of the euclidean norm. We also experimented with the unbounded euclidean distance for $\text{d}$, as done in \cite{crossmodaldist} but found that it leads to an unstable training. 

\subsubsection{Diversity Loss.}
We obtain diversity by enforcing the feature representation for both modalities to be distant across pairs with respect to the same distance $\text{d}$ as before. This spreads the features of different pairs apart in feature space. Due to the cross-modal loss these features mostly encode cross-modal information, thus ensuring sensitive feature representations for this content. The distance across pairs therefore contributes negatively into the loss
\begin{equation}
\label{eq:loss_diversity}
L_{div}(\mathbf{x}_i,\mathbf{y}_i,\mathbf{x}_j,\mathbf{y}_j) = -\frac{1}{2}\left[\text{d}(f(\mathbf{x}_i), f(\mathbf{x}_j)) +\text{d}(g(\mathbf{y}_i), g(\mathbf{y}_j))\right].
\end{equation}     
Here we assume our data to be diverse in the sense that the probability of the two pairs being very similar to each other is sufficiently low. This assumption holds true for most real world data and for modern datasets as they emphasize diversity in order to be challenging to solve. Nevertheless our approach handles two very similar pairs well since our loss is bounded limiting the impact of false information during training.

\subsubsection{Combining Both Loss Contributions.}
Our final training objective is given by a weighted sum of both loss contributions which leaves the weighting as an additional hyperparameter of our method. Determining this hyperparameter might sound like a challenging problem but we find that there are not many reasonable options in practice. We make two observations considering the balance of cross-modal and diversity loss. 1) Weighting $L_{cross}$ over $L_{div}$ encourages $f$ and $g$ to collapse their feature representation onto a single point. 2) Weighting $L_{div}$ over $L_{cross}$ significantly slows down the convergence of training and leads to a worse feature representation. It is important to note that these observations only hold true if the distance distributions in both loss contributions can quickly be equalized by the network during training. This is usually the case for modern deep CNN architectures with multiple normalization layers. 

Given our observations, we weight both loss contributions equally which yields our final loss
\begin{equation}
\label{eq:loss_combined}
L(\mathbf{x}_i,\mathbf{y}_i,\mathbf{x}_j,\mathbf{y}_j) = L_{cross}(\mathbf{x}_i,\mathbf{y}_i,\mathbf{x}_j,\mathbf{y}_j) + 
L_{div}(\mathbf{x}_i,\mathbf{y}_i,\mathbf{x}_j,\mathbf{y}_j).
\end{equation}
We obtain our training signal by randomly sampling $B$ tuples ($2B$ pairs) from our dataset and averaging the loss in equation \ref{eq:loss_combined} across the tuples. In section \ref{sec:ablation} we further validate our design choices.        

\section{Experiments}
In this section we quantitatively and qualitatively evaluate our self-supervised cross-modal pretraining mainly using RGB and optical flow as our modalities of choice. We will therefore refer to $f$ and $g$ as \textit{RGB-network} and \textit{OF-network} respectively. First, we present the implementation details for the self-supervised training. Afterwards, we quantitatively evaluate our feature representation by fine-tuning onto highly contested benchmarks and show a qualitative evaluation of the learned feature representation. Finally, we close with an extensive ablation study where we validate our design choices. 

\subsection{Implementation Details}
\label{sec:implementation}
We use the CaffeNet \cite{caffe,alexnet} architecture as this is well established for representation learning and extract our feature representation from fc6 after the ReLU activation. We use dropout with $p=0.5$. If not stated otherwise, we use batch normalization \cite{batchnorm} on all convolutional layers in order to be comparable with \cite{opn}. We conduct our experiments using pytorch \cite{pytorch}.

We use the trainset of UCF-101 for our pretraining in order to be consistent with previous work, the (TV-$L_1$) optical flow data is pre-computed using the OpenCV \cite{opencv} implementation with default parameters. Due to having temporally ordered data we use a stack of 10 consecutive optical flow frames and its temporally centered RGB frame for each pair. This is motivated by past work \cite{twostream14,posefromaction} which show that additional optical flow frames are beneficial for motion understanding. The input of the OF-network therefore has 20 channels. 

During training we randomly sample 60 videos ($B=30$ tuples) and from each video we randomly sample a pair. Similarly to \cite{opn,shufflealearn} the latter sampling is weighted by the average flow magnitudes in a pair which reduces the number of training samples without meaningful information.

The following augmentation scheme is applied independently to both pairs. Given the optical flow stack and the RGB frame we first rescale them to a height of 256 conserving the aspect ratio. We then randomly crop a 224x224 patch and apply random horizontal flips, both of which are done identically across the 10 flow frames and the RGB frame. We further augment the flow stack by applying random temporal flips (augmentation from \cite{posefromaction}) and subtract the mean \cite{twostream14}. Additionally we apply channel splitting \cite{opn} which yields a color-agnostic input forcing the network to learn a less appearance based representation. 

We find that a proper choice for $\epsilon$ is crucial for the stability of our training and recommend using $ \epsilon=1e-5 $ or bigger. We use SGD with momentum of 0.9 and weight decay of $5e-4$ with an initial learning rate of 0.01 which is reduced to 0.001 after 50K iterations. Training is stopped after 65K iterations and requires only \textit{6h GPU time} on a NVIDIA Titan X (Pascal).

\subsection{Action Recognition}
\label{sec:actionrec}
We evaluate our approach on the action recognition datasets UCF-101 and HMDB-51 which emerged as highly contested benchmarks for unsupervised representation learning on videos. To show the versatility of our approach we also utilize the considerably deeper VGG16 architecture, using its fc6 layer as our feature representation. We train on a single GPU using $B=12$ and a learning rate of $0.005$ for a total of 160K iterations, leaving all other hyperparameters unchanged. Pretraining on a single GPU completes in less than two days.

\subsubsection{Datasets.} 
The action recognition datasets are typically divided in three splits. UCF-101 is composed of 13K video clips per split, 9.5K for training and 3.5K for testing, divided in 101 action categories. HMDB-51 consists of 51 classes with 3.5K videos for training and 1.4K for testing per split. Although both datasets are created in a similar manner, HMDB-51 is more challenging as there can be multiple samples from the same video with different semantic labels (draw sword, sword exercise, sword fight). Samples from different classes can therefore have very similar appearance or posture information and require an understanding of motion to be distinguished properly \cite{hmdb}.	

\subsubsection{Finetuning Protocol.}  
We start with a randomly initialized CaffeNet and use the parameters of our pretrained RGB-network to initialize the convolutional and batch normalization layers (FC-layers start from scratch). For training and testing we follow the common finetuning protocol of \cite{twostream14} in order to be comparable with previous work \cite{shufflealearn,opn}. We provide test classification accuracies over 3 splits for both datasets. In order to ensure that the test set of UCF-101 does not contain images which have been previously seen during pretraining we conduct pretraining and finetuning on the same split. 

\subsubsection{Results.} 
\begin{table}
	\caption{Test classification accuracies over 3 splits for different pretraining schemes. If not stated otherwise the referenced methods use the CaffeNet architecture. +: indicates only Split 1. Our RGB-network outperforms previous state-of-the-art self-supervised methods especially on HMDB-51, while also being very efficient}
	\centering
	\setlength{\tabcolsep}{4pt}
	\begin{tabular}{lcccc}
		\hline\noalign{\smallskip}
		& Dataset & Traintime & UCF-101 & HMDB-51\\
		\noalign{\smallskip}
		\hline
		\noalign{\smallskip}
		Random & None & None & 48.2 & 19.5 \\
		ImageNet \cite{opn} & ImageNet & 3 days & 67.7 & 28.0 \\
		\hline
		Shuffle and Learn \cite{shufflealearn}  & UCF-101 & - & 50.2 & 18.1\\
		VGAN \cite{gvsd} (C3D) & flickr (2M videos) &  $>2$ days & 52.1 & - \\
		LT-Motion \cite{ltmotion} (RNN) & NTU (57K videos) & - & 53.0 & - \\
		Pose f. Action \cite{posefromaction} (VGG) & UCF,HMDB,ACT & - & 55.0 & 23.6 \\
		OPN \cite{opn} & UCF-101 & 40 hours & 56.3 & 22.1\\
		\hline
		Our & UCF-101 & 6 hours & \textbf{58.7} & \textbf{27.2} \\ 
		\hline
		\hline
		Random (VGG16)+ & None & None & 59.6 & 24.3 \\
		Our (VGG16)+ & UCF-101 & 1.5 days & 70.5 & 33.0 \\
		\hline
	\end{tabular}
	\label{table:actionrecognition}
\end{table}
As shown in Table \ref{table:actionrecognition}, our approach outperforms the previous state-of-the-art method \cite{opn} by over 2\% and 5\% on UCF-101 and HMDB-51 respectively. For the latter dataset our network almost closed the gap to its ImageNet pretrained counterpart. As mentioned before HMDB-51 requires a better understanding of motion and is not easily solved by appearance information. This favors our method because solving our pretraining task requires a good understanding of motion. Compared to past approaches which make use of optical flow \cite{posefromaction,ltmotion} our method exploits motion information much more effectively. We further expand this point in our qualitative evaluation and ablation studies.    

\subsection{Transfer Learning}
In this section, we evaluate the transferability of our learned feature representation to object classification and detection. To this end we use the Pascal VOC 2007 benchmark and compare to other recent unsupervised learning methods. 

\subsubsection{Datasets.} 
The Pascal VOC 2007 dataset \cite{pascal} provides annotation for 20 classes on 10K images, containing in total 25K annotated objects. The images are split equally between training/validation and testing set. The annotations provide class label and bounding box location for every object.

The ACT \cite{act} dataset is composed of 11K video clips of people performing various actions. The dataset contains 43 different action categories which can be summarized to 16 super-classes. For our pretraining we use the trainset containing 7K videos and do not make use of the semantic labels.  

\subsubsection{Pretraining and Finetuning Protocol.}
We apply our model onto the training videos from UCF-101, HMDB-51 and ACT \cite{act} which yield about 20K videos during pretraining. We double the number of iterations to 130K (dropping learning rate at 100K) and leave all other hyperparameters unchanged. Similar to section \ref{sec:actionrec} we use the Convolutional layers of our RGB-network to initialize a new CaffeNet for finetuning but we do not transfer the batch normalization layers, as suggested by \cite{opn}. In order to be comparable to \cite{opn} we finetune our RGB-network on Pascal VOC for the multi-class classification task following Kr\"ahenb\"uhl et al. \cite{krahenbuhl} protocol and use Fast-RCNN \cite{fastrcnn} framework when fine-tuning for object detection.

\subsubsection{Results.}
\begin{table}
	\caption{Results for Pascal VOC 2007 classification and detection. First segment shows baseline performance for supervised training. Methods in the second segment used large scale image data to pretrain their models whereas methods in the last segment used video data during pretraining}
	\centering
	\setlength{\tabcolsep}{4pt}
	\begin{tabular}{lcccc}
		\hline\noalign{\smallskip}
		& Dataset & Traintime & Classification & Detection \\
		\noalign{\smallskip}
		\hline
		\noalign{\smallskip}
		ImageNet \cite{opn} & ImageNet & 3 days & 78.2 & 56.8 \\
		\hline
		Context \cite{contextpred}  & ImageNet & 4 weeks & 55.3 & 46.6 \\
		Counting \cite{counting}  & ImageNet & - & 67.7 & 51.4 \\
		Jigsaw \cite{jigsaw}  & ImageNet & 2.5 days & 67.6 & 53.2\\
		Jigsaw++ \cite{jigsaw++}  & ImageNet & - & 72.5 & 56.5\\
		\hline
		Shuffle and Learn & UCF-101 & - & 54.3 & 39.9 \\
		OPN \cite{opn} & UCF,HMDB,ACT &  $>2$ days & 63.8 & 46.9\\
		Our & UCF,HMDB,ACT & 12 hours & \textbf{70.7} & \textbf{48.1} \\
		\hline
	\end{tabular}
	\label{table:pascal}
\end{table} 
Table \ref{table:pascal} shows a comparison of our approach to other unsupervised learning methods. Among the methods which do not make use of ImageNet data we improve upon state-of-the-art in both categories. We also show competitive accuracy in object classification among methods which do use ImageNet data. This result is remarkable as the datasets we used for pretraining are very dissimilar to the ImageNet or Pascal VOC dataset. This demonstrates the transferability of the feature representation obtained by our method. 

\subsection{Qualitative Evaluation}
\begin{figure}
	\centering
	\includegraphics[width=\linewidth]{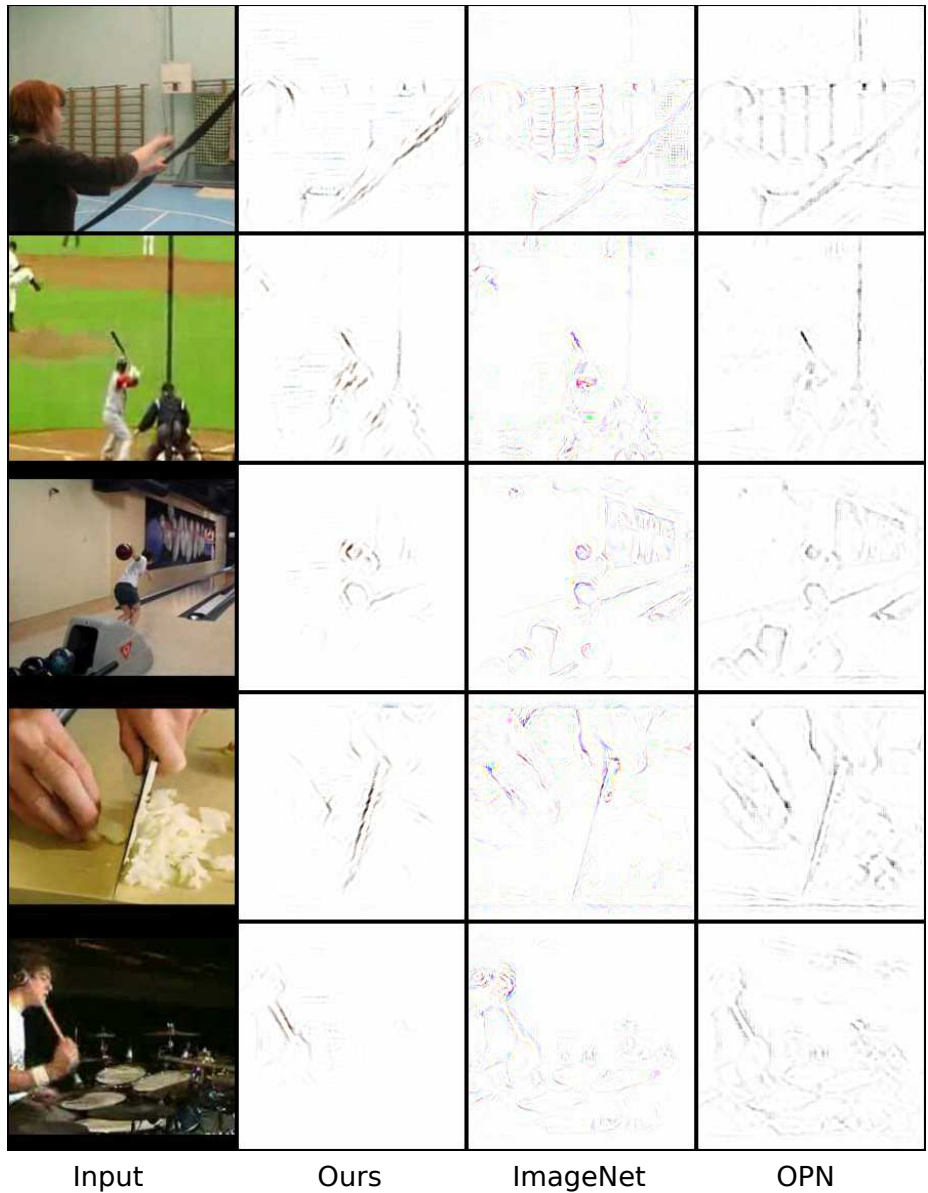}
	\caption{Comparison of the input activations for different models. The first column represents the input sample, second is our RGB-network, third is an ImageNet pretrained network and lastly OPN \cite{opn}. The activation are computed using guided backpropagation \cite{saliency}. In contrast to the other approaches our network is able to identify meaningful objects (Bow, bowling ball, drumstick ...) while also ignoring irrelevant background content}
	\label{fig:activations_1}
\end{figure}
\begin{figure}
	\centering
	\includegraphics[width=\linewidth]{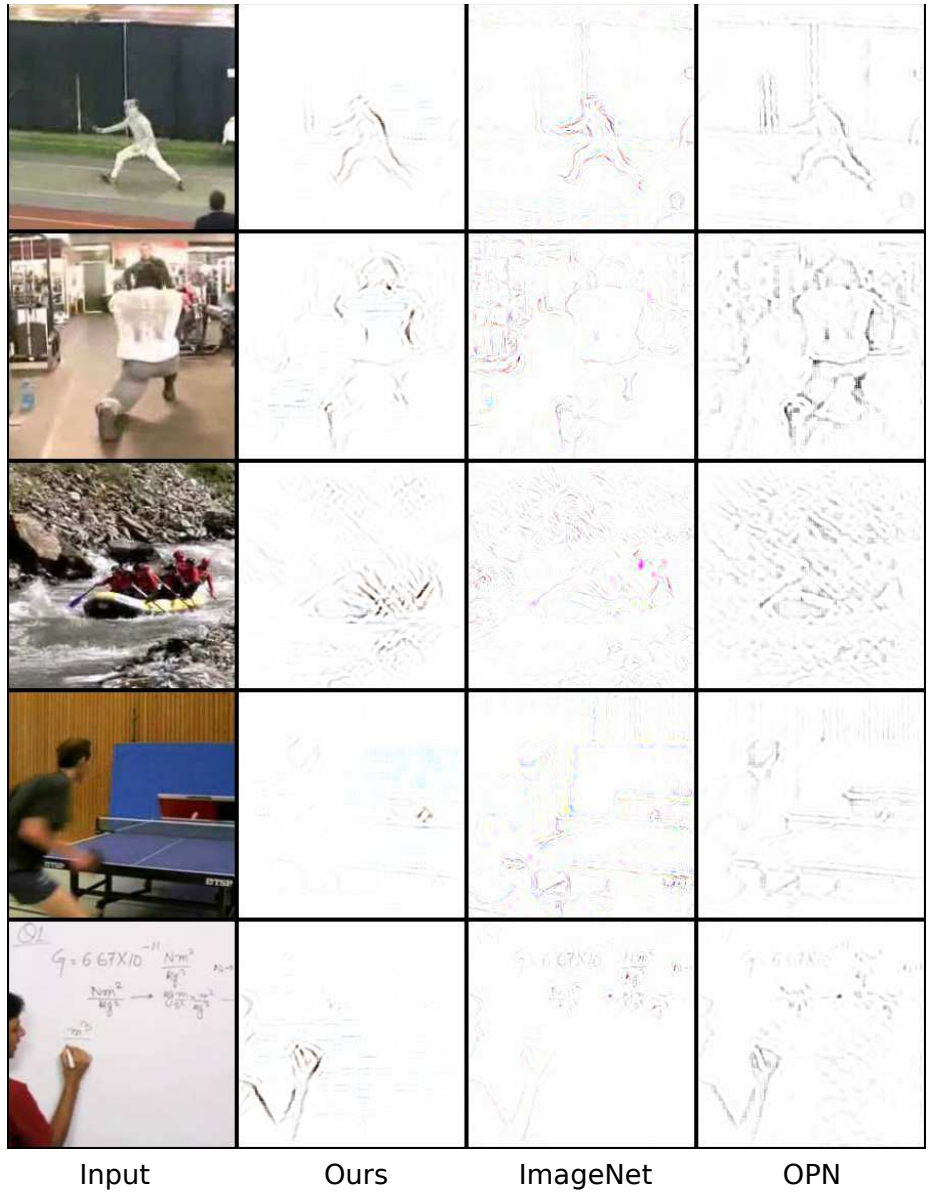}
	\caption{Additional input activations for our RGB-network, ImageNet pretraining and OPN \cite{opn}. Our network is also capable of identifying small moving objects (table tennis ball, pen), and ignore most of the background}
	\label{fig:activations_2}
\end{figure}
In this section we present a qualitative evaluation of our learned feature representations by showing input gradients obtained from high-level layers. We calculate the input layer activations for our pretrained RGB-network and compare to other approaches on the UCF-101 testset. To obtain the results in Fig. \ref{fig:activations_1} and \ref{fig:activations_2} we use guided backpropagation \cite{saliency} on the 100 strongest activations in pool5 of the networks. We can see that our training leads the network to focus on key moving objects involved in an action (bowling ball, drumstick, pen...) while ignoring irrelevant background information. Contrary, the ImageNet and OPN pretrained networks take the entire appearance of the scene into account indicating a more appearance based feature representation.

\subsection{Ablation Study}
\label{sec:ablation}
In this section we give a complete ablation study on all of our design choices. We start by investigating the effectiveness of our model without using optical flow as second modality. Afterwards we reveal that our method is mutually beneficial for the used modalities. We close by demonstrating the importance of our cross-modal loss and show that a more mainstream model design yields inferior results. We conduct our evaluation with CaffeNet on split 1 of UCF-101 and HMDB-51 only using UCF-101 for pretraining.

\subsubsection{Using Stack of Frame Differences as Second Modality.}
\begin{table}
	\caption{Comparing different choices of modalities. The left segment shows test accuracies obtained by using RGB and optical flow as modalities whereas the right segment reports test accuracies using RGB and SoD. We observe that all modalities benefit significantly from our pretraining}
	\centering
	\setlength{\tabcolsep}{4pt}
	\begin{tabular}{l|l|cc||cc}
		\hline\noalign{\smallskip}
		Dataset & Pretraining & RGB & OF & RGB & SOD \\
		\hline
		UCF-101 & Random weights & 49.1 & 76.4 & 49.1 & 64.5\\
		UCF-101 & Our & \textbf{59.3} & \textbf{79.2} & \textbf{55.4} & \textbf{66.3} \\
		\hline
		HMDB-51 & Random weights & 19.2 & 47.1 & 19.2 & 30.8 \\
		HMDB-51 & Our & \textbf{27.7} & \textbf{51.7} & \textbf{23.5} & \textbf{33.3} \\
		\hline
	\end{tabular}
	\label{table:sod_modalities}
\end{table}
In the works of \cite{oddoneout} and \cite{opn} reliable results were achieved using stack of frame differences (SOD) for action recognition. Inspired by their findings we train our model using RGB and SOD as modalities (we use the term SOD-network for $g$ in this context). We apply the same data sampling procedure as described in section \ref{sec:implementation} with slight modifications. To obtain a SOD we sample 5 RGB frames and calculate the difference between successive frames yielding us a SOD with 4 frame differences. The RGB input is then just the temporally centered RGB frame. We only apply random crops and random horizontal flips to SOD but keep the same augmentation as \ref{sec:implementation} for RGB. All other hyperparameters are kept the same for pretraining and finetuning. 

Test classification results on split 1 of UCF-101 and HMDB-51 are reported in the right segment of Table \ref{table:sod_modalities}. The RGB-network achieves $55.4\%$ and $23.5\%$ test classification accuracy respectively which is still competitive to previous state of the art methods.

\subsubsection{Mutual Benefit Between the Modalities.}
In order to demonstrate the mutual benefit of our method we finetune our obtained OF and SOD-network onto split 1 of UCF-101 and HMDB-51. We apply the same scheme already used for RGB to finetune both networks. We compare to randomly initialized networks and report results in Table \ref{table:sod_modalities}. 

We observe that our pretraining increases the classification accuracy on all used modalities. The OF-network acts supervisory on the RGB-network and also vice versa. This result is quite remarkable given the huge performance gap on the classification task and also holds true for RGB and SOD. Although the weaker modality benefits more from the pretraining it does not drag down the results of the stronger modality but instead gives a considerable boost especially for HMDB-51.

\subsubsection{Alternative Design Choices.}
\begin{table}
	\caption{Ablation study for alternative design choices conducted on split 1 of UCF-101 and HMDB-51. Models were pretrained onto UCF-101. Utilizing $L_{div}$ only performs on par with random weights. A more mainstream model design leads to a worse performance}
	\centering
	\setlength{\tabcolsep}{4pt}
	\begin{tabular}{l|l|cc}
		\hline\noalign{\smallskip}
		Dataset & Pretraining & RGB & OF  \\
		\hline
		UCF-101 & Random weights & 49.1 & 76.4 \\
		UCF-101 & Only $L_{div}$ & 49.0 & 75.3\\
		UCF-101 & Concat & 57.6 & 77.8\\
		UCF-101 & Our & \textbf{59.3} & \textbf{79.2} \\
		\hline
		HMDB-51 & Random weights & 19.2 & 47.1  \\
		HMDB-51 & Only $L_{div}$ & 18.0 & 47.2  \\
		HMDB-51 & Concat & 24.5 & 49.9  \\
		HMDB-51 & Our & \textbf{27.7} & \textbf{51.7} \\
		\hline
	\end{tabular}
	\label{table:design}
\end{table}	
As already indicated in section \ref{sec:method} we obtain trivial feature representations if we do not utilize the diversity loss. In the following we also demonstrate the importance of our cross-modal loss. To this end we pretrain our model the same way as before but set the cross-modal loss contribution to zero. We then finetune its RGB-network on UCF-101 and HMDB-51 and present our findings in Table \ref{table:design}. Without the cross-modal loss we perform on par with a randomly initialized network which indicates a failed pretraining and highlights the importance of the cross-modal loss. 

To validate our design choices further we now compare to a more mainstream approach of exploiting cross-modal information which is inspired by the work of \cite{posefromaction}. We concatenate the fc6 layers of our RGB and OF-network (8192 channels) and put a modified CaffeNet binary classifier (fc7, fc8) on top of it. We preserve our sampling and augmentation protocol described in \ref{sec:implementation} in order to be as comparable to our approach as possible. Given two pairs $\mathbf{x}_i, \mathbf{y}_i$ and $\mathbf{x}_j, \mathbf{y}_j$ we now exploit the mutual information by posing a binary classification surrogate task. We assign the combinations $\mathbf{x}_i, \mathbf{y}_i$ and $\mathbf{x}_j, \mathbf{y}_j$ positive labels whereas $\mathbf{x}_i, \mathbf{y}_j$ and $\mathbf{x}_j, \mathbf{y}_i$ obtain negative labels. Essentially the network has to figure out if a presented RGB frame and flow stack belong to the same pair or not. We pretrain this model using the Cross-Entropy loss and use the same hyperparameters as our original model. Finetuning is then performed for both modalities in the same way as before. We reference this alternative model as Concat. 

Results for our Concat model can be found in Table \ref{table:design}. We observe that the RGB-network of the Concat model is outperforming its randomly initialized counterpart but achieves significantly worse results than our pretraining. 

\section{Conclusion}
Recently self-supervised learning emerged as an attractive research area since unannotated data is available on a large scale whereas annotations are expensive to obtain. In our work we reveal cross-modal information as an alternative source of supervision and present an efficient method to exploit this signal in order to obtain powerful feature representations for both modalities. We evaluate our method in the context of action recognition where we significantly outperform previous state-of-the-art results on the two highly contested action recognition datasets UCF-101 and HMDB-51. We also show the transferability of our feature representations by achieving competitive results in object classification and detection. We close out by conducting an extensive ablation study in which we validate all our design choices. 

\section*{Acknowledgments}
We are grateful to the NVIDIA corporation for supporting our research, the experiments in this
paper were performed on a donated Titan X (Pascal) GPU.
\bibliographystyle{splncs04}
\bibliography{mybibliography}
\end{document}